\documentclass{article}
\usepackage{spconf,amsmath,graphicx} 
\usepackage[textwidth=0.7in]{todonotes}
\usepackage{tabularray}
\usepackage{multicol}
\usepackage{tikz}
\usepackage{float}
\usepackage[colorlinks=true,allcolors=black]{hyperref}
\usepackage{breakurl}

\title{ On the Importance of Signer Overlap for\\ Sign Language Detection}
\name{Abhilash Pal, Stephan Huber, Cyrine Chaabani, Alessandro Manzotti, Oscar Koller}
\address{\small Microsoft, Germany\\
\texttt{\footnotesize [t-apal,stephanhuber,t-cchaabani,amanzotti,oskoller]@microsoft.com}}
\begin{document}
%
\maketitle
\vspace{-2mm}
\begin{abstract}
Sign language detection, identifying if someone is signing or not, is becoming crucially important for its applications in remote conferencing software and for selecting useful sign data for training sign language recognition or translation tasks. We argue that the current benchmark data sets for sign language detection estimate overly positive results that do not generalize well due to signer overlap between train and test partitions. We quantify this with a detailed analysis of the effect of signer overlap on current sign detection benchmark data sets. Comparing accuracy with and without overlap on the DGS corpus and Signing in the Wild, we observed a relative decrease in accuracy of 4.17\% and 6.27\%, respectively.
Furthermore, we propose new data set partitions that are free of overlap and allow for more realistic performance assessment.  We hope this work will contribute to improving the accuracy and generalization of sign language detection systems.
\end{abstract}
\vspace{-2mm} 

\section{Introduction}
\label{sec:intro}
Sign languages, the natural languages of the deaf, use a variety of body articulators to convey meaning. The hands (hand shape, orientation, movement, and place of articulation), facial expression (eyebrows, eye gaze, mouth, cheeks, and head pose) and upper body pose work together to form meaningful sequences of signs. The visual characteristic of signed languages naturally excludes them from many computer interfaces that only support spoken languages.
With the advances of remote and hybrid work people frequently use remote conferencing software in everyday life. Voice activity detection allows speakers to be highlighted on the remote stage. This helps the audience to focus on the active speaker. However, in such a setup, sign language users remain hidden when they start to sign. The feasibility of remote schooling with a larger group of sign language users depends on the availability of sign language detection, a binary classification task that distinguishes signing from non-signing.
Yet, there are more applications for sign language detection. Sign language data is generally scarce. Sign language detection can help to identify video data that covers signing in pools of mixed signed and spoken content. 

At a first glance, sign language detection appears to be an established research field receiving regular contributions~\cite{cherniavsky_activity_2008, cherniavsky_activity_2009,santemiz_automatic_2009,khan_pause_2014,shipman_speed-accuracy_2017,borg_sign_2019,farag_learning_2019,monteiro_tradeoffs_2019, moryossef_realtime_2020,ismail_arabic_2021}, building on a selection of public benchmark corpora~\cite{borg_sign_2019,moryossef_realtime_2020}.
However, in the scope of this paper, we argue that current benchmark data sets estimate overly positive results that do not generalize well as they contain signer overlap between train and test partitions.
We make the following contributions: 
\begin{enumerate}
    \item \vspace{-1mm} We measure the effect of signer overlap on current sign detection benchmark data sets DGS Corpus and Signing in the Wild.
    \item \vspace{-1mm} We propose new data set partitions that are free of overlap and hence allow for more realistic performance assessment~\footnote{\url{https://github.com/oskoller/controlling-signer-overlap-for-sign-language-detection} \label{our_repo}}.
\end{enumerate}
 
In the remainder of this paper, we first describe the state of the art in Section~\ref{sec:relatedwork}, then introduce the employed data sets in Section~\ref{sec:datasets}, describe our method of signer identification by clustering for one data set in Section~\ref{sec:signerclustering} and show the effect of signer overlap between train and test partitions on sign detection in Section~\ref{sec:signdetection}. Finally, we will end with conclusion and future work in Section~\ref{sec:conclusionandfuturework}.
\vspace{-3mm} 
\section{Related Work}
\label{sec:relatedwork}
\vspace{-1mm}

Computational sign language processing has been gaining a lot of traction over the past years~\cite{camgoz_neural_2018,bragg_sign_2019,yin_including_2021,muller_findings_2022,saunders_signing_2022}. Next to sign language recognition, translation and production,  sign language detection remains an important and unsolved area: the classification of video sequences as containing a signing person or not. Most of the work on sign language detection focuses on 2-step systems: a feature extraction component followed by a classifier. 

For instance, Borg et al.~\cite{borg_sign_2019} resorts to a CNN pre-trained on ImageNet as a feature extractor. The obtained frame-based string features are then piped to an RNN architecture for classification. This approach improved the state of the art and reaches a classification accuracy of 87.67\%, hence outperforming the state of the art by approximately 18\%. The experiment is conducted on the Signing in the Wild data set.

 Another major contribution in the field of sign language detection is Moryossef et al.~\cite{moryossef_realtime_2020}. They use OpenPose~\cite{cao_openpose_2021} for full-body pose estimations, including 137 points per frame (70 for the face, 25 for the body, and 21 for each hand). Several experiments are then conducted in terms of classifiers: different linear baselines with a fixed context 
 and recurrent models with different counts of input features. \cite{moryossef_realtime_2020} introduces the DGS Corpus~\cite{hanke_extending_2020} for sign detection and achieves 91.53\% accuracy on the test set.
  
On a first glance, the mentioned classification results appear to be strong baselines. Upon careful inspection, however, we detected signer overlap between the train, dev and test partitions in both works. 
In fact, Borg et al.~\cite{borg_sign_2019} generate segments by stacking features of 20 video frames. 
This pool of segments is split into train, dev and test partition. This approach leads to an overlap in the videos and, hence, the signers across the different splits.  
Furthermore, after analysing the data splits used in Moryossef et al.~\cite{moryossef_realtime_2020}, we have come to an analogous conclusion: as a matter of fact, there is a signer overlap between the splits. This, we believe, could have contributed to the high accuracy reported by state-of-the-art systems. Our work focuses on analysing the effect of signer overlap on the accuracy for sign language detection tasks and proposes new data set splits that allow for more realistic performance evaluation.
\vspace{-3mm} 
\section{Data Sets}
\label{sec:datasets}

In the scope of this work, we analyse the two largest and most promising public sign detection data sets, namely DGS Corpus and Signing in the Wild.

\vspace{-3mm} 
\subsection{DGS Corpus}
\label{ssec:datasets_dgscorpus}
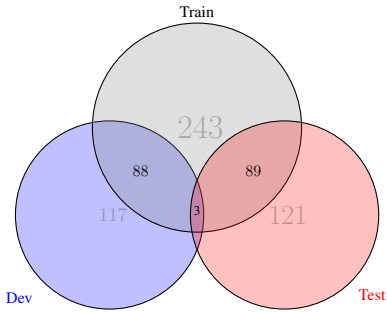
\begin{figure}[tbp]
        \centering
        \resizebox{0.6\linewidth}{!}{%
\begin{tikzpicture}[ 
    set/.style = {draw, circle,
        minimum size = 4.95cm}]
  
\node (A) [
    set,
    fill = blue,
    fill opacity = 0.25,
    text opacity = 1,
    label = {[label distance=0.2cm,blue]225:\large Dev},
    thick
    ] {\Large  \textbf{$117$}};
\node (B) at (45:3.25cm) [
    set,
    minimum size = 5.5cm,
    fill = gray,
    fill opacity = 0.25,
    text opacity = 1,
    label={[label distance=0.01cm]90:\large Train},
    thick
    ] {\Huge \bf $243$};
\node (C) at (0:4.6cm) [
    set,    
    fill = red,
    fill opacity = 0.25,
    text opacity = 1,
    label={[label distance=0.1cm,red]315:\large Test},
    thick
    ] {\huge $121$};
 
\node at (barycentric cs:A=1,B=1) [left] {\large $88$};
\node at (barycentric cs:B=1,C=1) [right] {\large $89$};
\node at (barycentric cs:A=1,B=0.1,C=1) [] {3};
\end{tikzpicture}%
}
        \caption{\small Number of signers per train, dev, and test corpus of~\cite{moryossef_realtime_2020}. Numbers in the overlapping regions indicate signer overlaps between the sets. \vspace{-4mm}}
        \label{fig:datasets_dgsoriginal_overlap}
    \end{figure}
DGS Corpus is a German sign language data set.
Its total footage contains over 1150 hours of recordings, of which only a part is annotated. 
The Public DGS Corpus~\cite{hanke_extending_2020} is a subset of about 50 hours of the DGS Corpus intended for public release. \cite{moryossef_realtime_2020} used it to propose a train, dev and test split for the task of sign language detection. 
Their proposed three splits cover a total of 623 videos, which are on average 9 minutes long and show the dialogue of two signers in front view. For the task of sign detection, the annotations covering a translation to spoken language are used to indicate when signing starts and ends.
Table~\ref{tab:corpusstats_dgs} shows the statics of the original splits as proposed by~\cite{moryossef_realtime_2020}.

In Section~\ref{sec:signerclustering} we introduce our approach to identify the signers present in train, dev and test splits and their overlap. We find that the same 88 signers occur both in train and dev, while 89 signers overlap between train and dev. This is summarized in Figure~\ref{fig:datasets_dgsoriginal_overlap}. To assess the effect of this, we take the signer information and split the original test set into a part with train overlap and a part with no overlap, as can be seen in Table~\ref{tab:corpusstats_dgs_test_split}. Finally, with this work, we also propose and release\textsuperscript{\ref{our_repo}} a new data split with no overlap among train, dev and test sets. Each signer appears in only one split, and we used a 60:20:20 ratio, compare Table~\ref{tab:corpusstats_dgs_proposed}. We claim that this set is more suitable for benchmarking.

\begin{table}[tbp]
    \centering
    \begin{tabular}{c|ccc}
 &	Train &	Dev	& Test-Original \\\hline
Hours & 48.90 & 22.92 & 24.33\\
Number of Signers & 243 & 117 & 121\\
    \end{tabular}

    \caption{ \small Showing corpus statistics of the original DGS Corpus split proposed by \cite{moryossef_realtime_2020}. The number of signers is estimated based on the proposal in Section~\ref{sec:signerclustering}.}
    \label{tab:corpusstats_dgs}
\end{table}
\begin{table}[tbp]
\setlength{\tabcolsep}{2pt} 
    \centering
    \begin{tblr}{rowsep=0.06pt,colspec={Q[c,m] | Q[c,m] | Q[c,m] Q[c,m]}}
         & Test-Original & {Test with\\Train Overlap} & {Test no\\Train Overlap}  \\\hline
Hours &  24.33 & 16.09 & 7.23\\
Signers &  121 & 89 & 32\\
    \end{tblr}
    \caption{\small Showing statistics of splitting DGS Corpus original test set into a part with signer overlap and a part without signer overlap.}
    \label{tab:corpusstats_dgs_test_split}
\end{table}
\begin{table}[tbp]
    \vspace{-4mm}
    \centering
    \begin{tabular}{c|ccc}
    & \multicolumn{3}{c}{Proposed} \\
 &	Train &	Dev	& Test \\\hline
Hours & 45.93 & 15.33 & 15.32\\
Number of Signers & 183 & 64 & 62\\
    \end{tabular}
    \caption{ \small Showing corpus statistics of the proposed DGS Corpus split with no signer overlap between the sets.}
    \label{tab:corpusstats_dgs_proposed}
\end{table}



\vspace{-2mm} 
\subsection{Signing In the Wild}
\label{ssec:datasets_signinginthewild}
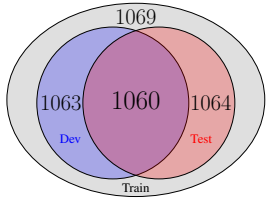
\begin{figure}[tbp]
        \centering
        \resizebox{0.4\linewidth}{!}{%
\begin{tikzpicture}[ 
    set/.style = {draw, circle,
        minimum size = 4.95cm}]
\draw (0.75cm,0.1cm) ellipse (4.2cm and 3.15cm) [
    set,
    fill = gray,
    fill opacity = 0.25,
    text opacity = 1,
    thick
    ] node (TRAIN) {};
\node (DEV) [
    set,
    fill = blue,
    fill opacity = 0.25,
    text opacity = 1,
    label = {[label distance=-1.2cm,blue]225:\large Dev},
    thick
    ] {};
\node (TEST) at (0:1.5cm) [
    set,    
    fill = red,
    fill opacity = 0.25,
    text opacity = 1,
    label={[label distance=-1.2cm,red]315:\large Test},
    thick
    ] {};
 
\node at (DEV.center) [left,xshift=-0.9cm] {\huge $1063$};
\node at (TRAIN.center) {\Huge $1060$};
\node at (TRAIN.center) [below, yshift=-2.6cm]{\large Train};
\node at (TRAIN.center) [above, yshift=2.35cm]{\huge $1069$};
\node at (TEST.center) [right,xshift=0.9cm] {\huge $1064$};
\end{tikzpicture}%
}
        \caption{\small Number of videos per train, dev, and test corpus of Signing in the Wild, splitting as proposed by~\cite{borg_sign_2019}. Numbers in the overlapping regions indicate video/signer overlaps between the sets. \vspace{-4mm}}
        \label{fig:datasets_signinginthewildoriginal_overlap}
    \end{figure}

The Signing in the Wild data set is a collection of videos harvested from Youtube \cite{borg_sign_2019}. The authors' data selection was designed to incorporate as many sign languages, signers, and footage settings as possible. The data set also has two non-signing classes, namely “speaking” and “other”. 
The authors considered a temporal context of 10 frames surrounding the current frame to create the labels. Ambiguous cases were not labelled.
The data set is composed of 1120 videos in total, but given that some are not available on Youtube anymore, we are left with 1069.
In \cite{borg_sign_2019}, the frames are extracted at a constant rate of 25 fps and then resized to 224x224. At a later stage, sequences of 20 frames are joined and attributed a groundtruth label.
The original authors~\cite{borg_sign_2019} use 5-fold validation for testing. However, they shuffle and partition their data on the segment level, which allows sequences from the same video, and hence signer, to be in different folds. This overlap is quantified in Figure~\ref{fig:datasets_signinginthewildoriginal_overlap}.
To evaluate the effect of video/signer overlap between the different splits, we propose and release\textsuperscript{\ref{our_repo}} a different partitioning which ensures videos are not spread across different splits and hence there is no video/signer overlap. Refer to Table~\ref{tab:corpusstats_signinginthewild_ourapproach} for statistics.
\begin{table}[tbp]
    \centering
    \begin{tabular}{c|ccc}
 &	Train &	Dev	& Test \\\hline
Number of Videos & 856 & 108 & 107\\
Number of Segments & 113934 & 13826 & 14206\\
    \end{tabular}
    \caption{\small Showing corpus statistics of proposed Signing in the Wild split with no train overlap.}
\label{tab:corpusstats_signinginthewild_ourapproach}
\vspace{-2mm}
\end{table}

\vspace{-3mm} 
\section{Signer Clustering}
\label{sec:signerclustering}

Opposed to Signing in the Wild, DGS Corpus contains a number of different recordings where the same signers re-occur. However, no labelling of these signers is available. Hence, to control signer overlap in DGS Corpus, we need to identify which signer is present in which videos. In line with previous work~\cite{vaezi_joze_msasl_2019}, we applied face clustering to the data set. We opted for the density-based clustering algorithm DBSCAN (Density Based Spatial Clustering of Application with Noise) operating on extracted face embeddings.

To measure clustering performance, we manually labelled 11 female and 14 male signers occurring in 83 videos of the DGS Corpus with at least 2 and at most 7 videos per signer. The set allows to assess the quality of the resulting clustering by verifying that there is no overlap across the 25 signers. A signer is considered to be correctly clustered if all videos from the signer occur in only one class, and that class has not been attributed to a different signer, nor is it the garbage class. 
We measure clustering accuracy as the number of correctly clustered signers divided by the total number of labelled signers.

 Based on the (reasonable) assumption that the DGS Corpus contains a single signer per video only, we sampled a fixed amount of gallery images from each video, all representing the same signer. We then encoded them using the CNN model from the face\_recognition\footnote{\url{https://github.com/ageitgey/face_recognition}} library into 128 dimensional vectors and clustered them using the DBSCAN algorithm. The encoder model is a Maximum Margin CNN from the dlib library~\cite{king_dlibml_2009}. DBSCAN works by finding regions of high density that are separated by regions of low density. It labels points as core, border, or noise based on their density, and eventually forms clusters by connecting core points that are neighbors, and adding border points to the nearest cluster. The remaining frames which are deemed to be noise form the garbage class. 

Next, we fit a DBSCAN model to the extracted images, and each video is assigned into one single cluster based on majority voting over the frames extracted from the video and measure clustering performance on the manually labelled set.

With this as the training and testing paradigm, we do a grid search for the best number of gallery images per video. The results are summarized in Table~\ref{tab:signerclustering_galleryimages} reporting the maximum or peak accuracy for 1, 5 or 20 gallery images. We see that by increasing the gallery images the accuracy increases with a peak at 20 images. The accuracy is single peaked for the given epsilon, i.e., accuracy monotonically decreases if we increase or decrease the value of epsilon. 
In Table~\ref{tab:signerclustering_epsilon} we summarize the accuracy when varying epsilon.
The best result is with $epsilon = 0.36$ which correctly classifies all videos for 24 out of the 25 signers in our test set, and results in 308 unique signers detected. 

\begin{table}[tbp]
    \centering
    \begin{tabular}{l|ccc}
        Number of Gallery Images & 1 & 5 & 20 \\ \hline
        Test Set Accuracy [\%] & 24.0 & 92.0 & 96.0  \\ 
        Epsilon & 0.50 & 0.41 & 0.36  \\
    \end{tabular}
    \caption{\small Showing variance in the peak test set signer accuracy for different number of gallery images used. 
    }
    \label{tab:signerclustering_galleryimages}
\end{table}

\begin{table}[tbp]
    \centering
    \begin{tblr}{rowsep=0.1pt,colspec={Q[c,m] Q[c,m] Q[c,m] Q[c,m]}}
         Epsilon & {Number\\ of Signers} & {Videos in\\ Garbage Class} & Accuracy \\ \hline
         0.35 & 305 & 79 & 92.0  \\ 
         0.36 & 308 & 57 & 96.0  \\ 
         0.38 & 302 & 22 & 92.0  \\ 
         0.40 & 265 & 6 & 80.0  \\ 
         0.42 & 189 & 2 & 44.0  \\ 
    \end{tblr}
    \caption{\small Here, we are working with 20 gallery images per video. The number of videos in garbage class decreases as we increase epsilon since the size of neighbourhoods included as clusters increases and the videos assigned to noise decreases.}
    \label{tab:signerclustering_epsilon}
    \vspace{-2mm}
\end{table}

\section{Effect of Train-Test Signer Overlap for Sign Language Detection}
\label{sec:signdetection}

\subsection{DGS Corpus}
To model sign detection, we closely followed the approach proposed by~\cite{moryossef_realtime_2020}. As in their work, we first used OpenPose~\cite{cao_openpose_2021} to detect characteristic landmarks of full-body poses~\cite{schulderopenpose}. These include 137 points per frame: 70 for the face, 25 for the body, and 21 for each hand%
. However, in real-world applications, video resolution and distance from the camera can vary greatly, making normalization of landmarks necessary. To achieve this, we normalized the landmarks such that the mean distance between the left and right shoulders of each person is 1.

To focus on movements of the signer, we utilized the optical flow of these landmarks, i.e., the spatial difference of each landmark between two consecutive frames. The optical flow was then normalized by the frame rate to create frame-rate independent representations. To minimize the memory and latency footprint of the model, we employed a lightweight architecture consisting of a dropout layer to prevent overfitting to specific landmarks \& a single unidirectional LSTM with a hidden size of 64 followed by a linear classification head.

\begin{table}[tbp]
    \centering
    \begin{tblr}{rowsep=0.1pt,colspec={Q[c,m] | Q[c,m] Q[c,m]}}
         Test-Original & {Test with\\Train Overlap} & {Test no\\Train Overlap}  \\\hline
         87.81 &  89.00 & 85.29\\
    \end{tblr}
    \caption{\small Detection accuracy [\%] on different test partitions of DGS corpus. Test-Original is proposed by~\cite{moryossef_realtime_2020} and has overlap with signers in training. It is split into a part with and without train overlap. \vspace{-5.5mm}}
    \label{tab:results_detection_dgs}
\end{table}

To measure the impact of signer overlap, we investigate the performance of our system on the original test set as well as on its subsets with and without signer overlap. For our experiments we rely on the official implementation\footnote{\url{https://github.com/google-research/google-research/tree/master/sign_language_detection}}
of \cite{moryossef_realtime_2020}. The results are summarized in Table~\ref{tab:results_detection_dgs}. Despite taking great care and communicating with the original authors, our reproduced performance falls short of the reported one on the original test set. The accuracy on Test-With-Train-Overlap exceeds that on Test-No-Train-Overlap by 3.71 absolute percentage points. We attribute this margin to the fact that the model overly regresses to characteristic movement patterns of signers who are common to both train set and Test-With-Train-Overlap.
 
On our newly proposed sets with no signer overlap between the splits (refer to Section~\ref{ssec:datasets_dgscorpus} and Table~\ref{tab:corpusstats_dgs_proposed}), we achieve an accuracy of 84.9\% and 85.8\% on dev and test, respectively. This seems to be in line with the results on the divided test set (Test-no-train-overlap). We suggest to adopt the proposed set as it contains significantly more signers in the test set and hence allows for more robust and realistic performance evaluation. 
 
\subsection{Signing in the Wild}

To evaluate the effect of overlap of video segments on sign language detection systems for Signing in the Wild, we train the proposed system in \cite{borg_sign_2019} on two different sets of splits described in \ref{ssec:datasets_signinginthewild}.
Initially, we downloaded the videos contained in the data set from YouTube, then extracted features per frame with a VGG architecture pretrained on ImageNet. Once the input features were obtained, segments of length 20 (frames) were generated with a stride of 20.
Two different approaches were carried out to train and test the final RNN-architecture of the sign language detection system: one with video/signer overlap based on the splitting proposed by~\cite{borg_sign_2019} and another one without overlap based on our proposed splitting. 
Table~\ref{tab:corpusstats_signinginthewild_results} summarizes our findings. We see that the accuracy of the model with signer overlap exceeds the one with no overlap by 6.8 and 4.6 absolute percent points on the dev and test sets, respectively. Hence, controlling for video/signer overlap is crucial to get a realistic and generalizing performance estimation.
We also release this new split\textsuperscript{\ref{our_repo}} for Signing in the Wild as we suggest to adopt it for future benchmarking.
\begin{table}[tbp]
    \centering
    \begin{tabular}{c|ccc}
 &	Dev Accuracy	& Test Accuracy \\\hline
With signer overlap  & 91.4 & 89.3\\
No signer overlap  & 84.6 & 83.7\\
    \end{tabular}
    \caption{\small Detection accuracy in [\%] on Signing in the Wild~\cite{borg_sign_2019} using two different paradigms to data splitting.\vspace{-5mm}}
    \label{tab:corpusstats_signinginthewild_results}
\end{table}
\section{Conclusion and Future Work}
\label{sec:conclusionandfuturework}
Our results on current benchmark data sets of Signing in the Wild and DGS Corpus suggest that signer overlap impacts performance and, hence, does not allow for realistic performance assessment.
Controlling the distribution of signers, we propose and publicly release signer-independent splits for both data sets which we hope will contribute to improving the accuracy and generalization of sign language detection systems.
Controlling signer overlap is an important step to accommodate fairness, accountability, transparency, and ethics consideration for sign language data sets~\cite{bragg_fate_2021a}. Moreover, the proposed signer clustering could also help to address privacy concerns~\cite{bragg_exploring_2020}.
Of course, we did not achieve a perfect accuracy on signer identification and clustering. Particularly, for Signing in the Wild we assumed each video contains only a single signer.
Improving this and controlling for remaining overlap will be important for the future, especially for web-scraped videos. 

\bibliographystyle{IEEEbib}
\bibliography{oskoller,main} 

\end{document}